\pdfoutput=1

\documentclass[11pt]{article}

\usepackage[final]{emnlp2021}

\usepackage{times}
\usepackage{latexsym}

\usepackage[T1]{fontenc}

\usepackage[utf8]{inputenc}

\usepackage{microtype}

\usepackage{helvet} 
\usepackage{courier}  
\usepackage{url}  
\usepackage{graphicx} 
\usepackage[noend]{algpseudocode}
\usepackage{algorithmicx,algorithm}
\usepackage{graphicx}
\usepackage{amsmath}
\usepackage{booktabs}
\usepackage{booktabs}
\usepackage{enumitem}
\usepackage{float} 
\usepackage{subfigure}
\usepackage{setspace}
\usepackage{helvet}  
\usepackage{courier}  
\usepackage{xcolor}
\usepackage{color}
\usepackage{multicol}
\usepackage{multirow}
\usepackage{esvect}
\usepackage{epstopdf}
\usepackage{array}
\usepackage{amsmath,amsfonts,amssymb,amsthm, bm}
\usepackage{mathrsfs}
\usepackage{verbatim}
\usepackage{enumitem}
\usepackage{latexsym}
\usepackage{setspace}
\usepackage{amsmath}
\usepackage{amssymb}
\usepackage{mathrsfs}

\title{RoR: Read-over-Read \\ for Long Document Machine Reading Comprehension}

\author{Jing Zhao$^1$, Junwei Bao$^{1}$\thanks{Corresponding Author: {baojunwei001@gmail.com}}, Yifan Wang$^1$, Yongwei Zhou$^2$, \\ 
\bf Youzheng Wu$^1$, Xiaodong He$^1$, Bowen Zhou$^1$\\
  $^1$JD AI Research, Beijing, China\\
  $^2$Harbin Institute of Technology, Harbin, China \\
  \tt{\{zhaojing857,baojunwei,wangyifan15\}@jd.com} \\
    \tt{\{wuyouzheng1,xiaodong.he,bowen.zhou\}@jd.com} \\
      \tt {ywzhou@hit-mtlab.net} \\
  }

\date{}

\begin{document}
\maketitle

\begin{abstract}
Transformer-based pre-trained models, such as BERT, have achieved remarkable results on machine reading comprehension.
However, due to the constraint of encoding length (\textit{e.g.,} 512 WordPiece tokens), a long document is usually split into multiple chunks that are independently read.
It results in the reading field being limited to individual chunks without information collaboration for long document machine reading comprehension.
To address this problem, we propose \textbf{RoR}, a \textit{read-over-read} method, which expands the reading field from chunk to document.
Specifically, RoR includes a chunk reader and a document reader.
The former first predicts a set of regional answers for each chunk, which are then compacted into a highly-condensed version of the original document, guaranteeing to be encoded once.
The latter further predicts the global answers from this condensed document.
Eventually, a voting strategy is utilized to aggregate and rerank the regional and global answers for final prediction.
Extensive experiments on two benchmarks QuAC and TriviaQA demonstrate the effectiveness of RoR for long document reading. 
Notably, RoR ranks 1st place on the QuAC leaderboard \footnote{\url{https://quac.ai/}} at the time of submission (May 17th, 2021)\footnote{Our code is available at \url{https://github.com/JD-AI-Research-NLP/RoR}}.
\end{abstract}

\section{Introduction}
\noindent
The task of machine reading comprehension (MRC), which requires machines to answer questions through reading and understanding a given document, has been a growing research field in natural language understanding~\cite{hermann,newsqa,squad1,squad2,triviaqa,quac}.

Transformer-based pre-trained models have been widely proven to be effective in a range of natural language processing tasks, including the MRC task~\cite{BERT,roBERTa,xlnet,electra}.
Typically, these models consist of a stack of transformer blocks that only encode a length-limited sequence (\textit{e.g.,} 512).
However, the input sequences in some MRC tasks may exceed the length constraint.
For example, each instance in open-domain MRC usually consists of a collection of passages, such as TriviaQA~\cite{triviaqa}, one of the most popular open-domain MRC datasets, containing 6,589 tokens on average. 
In addition, for conversational MRC task, such as QuAC~\cite{quac}, existing methods incorporate conversation history by prepending the previous utterances to the current question, which is packed with the document into a length input (707 tokens on average).
%

%

To handle a long document that exceeds the
length constraint, a commonly used approach is to split a document into multiple individual chunks and then predict answers from each chunk separately.
The highest scoring span in these answers is selected as the final answer. 
This approach is straightforward but results in two problems: (1) the reading field is limited to the regional chunk instead of the complete document; and
%
(2) the scores of the answers are not comparable as they are not globally normalized over chunks.
%

%

%

To address these problems, we propose \textbf{RoR}, a \textit{read-over-read} pipeline, which is able to expand the reading field from chunk-level to document-level.
RoR contains a chunk reader and a document reader, both of which are based on the pre-trained model.
Specifically, the chunk reader first predicts the regional answers from each chunk.
These answers are then compacted into a new document through a minimum span coverage algorithm guaranteeing that its sequence length is shorter than the limitation (\textit{i.e.,} 512).
By this means, all regional answers can be normalized in one document.
This document serves as the highly-condensed version of the original document, which is further read by the document reader to predict a set of global answers.
As the chunk reader and global reader provide high confidence answers from different views, we fully leverage both of them for final answer prediction.
Specifically, after predicting the regional and global answer spans, a voting strategy is proposed and utilized to rerank them.
This voting strategy is based on the idea that a candidate regional or global answer span overlapped more with the others is more likely to be correct.

%
%

The contributions are summarized as follows:
\begin{itemize}[leftmargin=*]

\item We propose a \textit{read-over-read} pipeline containing an enhanced chunk reader and a document reader, which is able to solve the problem of long document reading limitation in existing models.
\item We propose a voting strategy to rerank the answers from  regional chunks and a condensed document,  overcoming the major drawback in aggregating the answers from different sources. 
\item Extensive experiments on long document benchmarks are conducted to verify the effectiveness of our model. Especially on the QuAC dataset, our model achieves state-of-the-art results over all evaluation metrics on the leaderboard.

\end{itemize}

\section{Related Work\label{Sec2}}
MRC is a fundamental task in natural language
understanding that aims to determine the correct answers to questions after reading a given passage~\cite{hermann,newsqa,squad1,squad2}.
%
The best performing models in various MRC tasks are commonly based on the pre-trained language models (PLMs) within the typical encoding
limit of 512 tokens.
%
However, the input sequence in some MRC tasks usually exceeds the length limit,  such as conversational MRC and open-domain MRC.
%

\textbf{Conversational MRC}, which extends the traditional single-turn MRC, requires the models to additionally understand the conversation history~\cite{coqa,quac,gao,flowqa,gupta} as dialog and conversational recommendation systems~\cite{lu-etal-2021-revcore}.
A straightforward but effective approach of modeling the history is to prepend the previous dialogs to the current question, which will compose a lengthy input sequence with the relatively long document~\cite{chunk}.

\textbf{Open-domain MRC} is a task
of answering questions using a large collection of
passages~\cite{triviaqa,searchqa,natural}. The main challenge of this task is that the sequence length of multiple passages relevant to each question far exceeds the length limit of 512 tokens. For example, documents
in TriviaQA~\cite{triviaqa} contain 6,589 tokens on average.

To enable the PLMs to encode long documents, a common approach is to chunk the document into overlapping chunks of length 512, then process each chunk separately, which inevitably causes the two problems aforementioned.
Another intuitive approach is to increase the encoding length of the PLMs.
For example, the recently proposed PLMs Longformer~\cite{Longformer} and Big bird~\cite{big}, specifically for long document modeling, have extended the encoding length from 512 to 4,096. 
However, their encoding length is fixed. The two problems caused by chunking still exist when encoding the sequences longer than 4,096. 
In contrast, our proposed model RoR is flexible which is able to encode sequences of arbitrary length.
Moreover, RoR is assembleable and its encoder can be replaced with any PLMs, such as BERT and Longformer.

Theoretically, hierarchical models can be adapted to long document MRC~\cite{han,multi,beyond}.
However, deploying the large transformer-based PLMs as the encoders of hierarchical models can be prohibitively costly.
Typically, hierarchical models parallelly encode the splitted chunks of a long document with multiple transformers, which requires extremely large GPU support. 
In contrast, RoR only needs to read a chunk at each encoding process, then gradually predict all answers from chunk to document. Therefore, RoR is able to deal with a long document without consuming too much computing resources and can be more widely used than hierarchical models.

\begin{figure*}
\centering
\includegraphics[width=15cm,height=7.5cm]{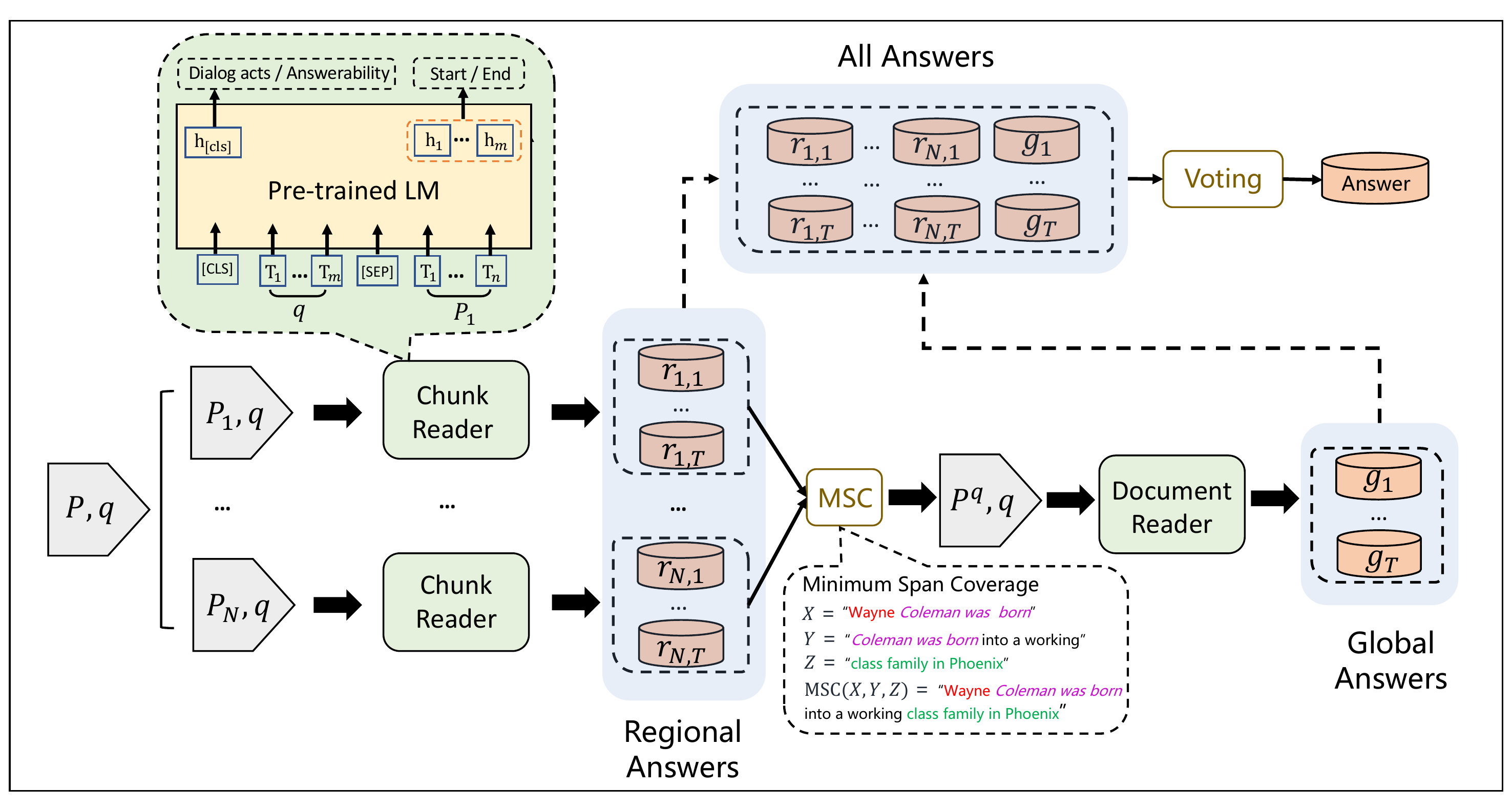}
\caption{The architecture of the proposed \textit{Read-over-Read} pipeline.}
\label{framework}
\end{figure*}
\section{Approach}
\subsection{Task Formulation}
Given a document $P$, a question $q$, the task of MRC is to predict an answer span $y$ from $P$ based on the comprehension of $P$ and $q$.
If $q$ is an unanswerable question, the QuAC dataset requires the model to give an unanswerable tag as the final answer.
To model the dialog history in QuAC, we prepend previous pairs of (question, answer)  to the current question to form the question $q$. 
Formally, $q=[H_k ; \mathtt{[SEP]} ; q_k]$, where $q_k$ is $\mbox{$k$-th}$ question and $H_k$ is dialog history.


Additionally, a special task in the QuAC dataset is dialog act prediction. QuAC provides two dialog acts, namely, continuation (Follow up) and affirmation (Yes/No).
The continuation dialog act consists of three possible labels (follow up, maybe follow up or don’t follow up). 
The affirmation dialog act also consists of three labels (yes, no or neither).
Both two dialog act predictions are three-label classification tasks.

\subsection{Framework Overview}
The architecture of our proposed \textit{read-over-read} pipeline is illustrated in Figure \ref{framework}. RoR includes a chunk reader and a document reader, both of which employ the PLM as the text encoder.
Given a data sample ($P$, $q$), we split it into multiple chunk-based sample \{$(P_1, q), ..., (P_N, q)$\} with slide-window, where $N$ is the number of split chunks.
The chunk reader first predicts a set of regional answers $\{\{r_{i,j}\}^T_{j\mbox{=}1}\}^N_{i\mbox{=}1}$ from all chunks, 
where $T$ is the max number of the predicted answers for one chunk.
The regional answers are then compacted to a new document $P^q$ by a minimum span coverage algorithm (MSC).
Notably, most of the answers in TriviaQA dataset are named entities that cannot reflect enough contextual information. 
Therefore, for the TriviaQA dataset, we use the sentences where the regional answers are located to compact to $P^q$.
$P^q$ is the condensed version of original document $P$ to the question $q$, which is packed into the text encoder at once.
$P^q$ is further read by the document reader to predict the global answers $\{g_i\}^T_{i\mbox{=}1}$.
The answers from all chunks and the condensed document are aggregated together and reranked by a voting strategy to choose the final answer.

\subsection{Chunk Reader}
The chunk reader predicts the regional answers for each chunk based on the contextualized representations of a given document which are obtained by the pre-trained encoder. This section introduces the components of the chunk reader in detail. 
\subsubsection{Text Encoder}
The goal of a text encoder is to convert the input sequence  into a series of contextualized feature representations $\{\mathbf{h}_i\}^L_{i=1}$, where $L$ is the length of input sequence.
The input sequence of the encoder contains a chunk $P$, the question $q$
which are concatenated to one token sequence with a special splitter $\mathtt{[SEP]}$, represented as $\mathtt{X} = [\mathtt{[CLS]} ; q ; \mathtt{[SEP]} ; P]$, where $\mathtt{X}$ is the input token sequence. the representation of $\mathtt{[CLS]}$ is treated as the sentence-level feature for the sentence classification tasks, \textit{i.e.} answerability and dialog acts prediction.
%

%
%

\subsubsection{Answer Prediction}
The answers prediction of conversational MRC requires two levels of feature, token level feature for predicting answer span and sentence level feature for predicting dialog acts and answerability. Open-domain MRC only requires token level feature.

\noindent \textbf{Token level answer}. The encoder representations $\{\mathbf{h}_i\}^L_{i=1}$ serve as the token level features, which are used to compute the probability of each token being the begin or the
end of an answer span.
Concretely, $\{\mathbf{h}_i\}^L_{i=1}$ are projected onto start logit and end logit through multi-layer perceptrons separately, which are  then sent to a softmax function to compute the start and end probability distributions along all tokens in this sequence.
The probabilities of each token being the begin and the
end of predicted span are calculated as follows:
\begin{align}
    r^s_i & =  \mathbf{W}^s_2\mathtt{tanh}(\mathbf{W}^s_1\mathbf{h}_i) \label{start}\\
    r^e_i & =  \mathbf{W}^e_2\mathtt{tanh}(\mathbf{W}^e_1[\mathbf{h}_i;\mathbf{h}_s]) \label{beam} \\
    p^s & =  \mathtt{softmax}(r^s) \\
    p^e & =  \mathtt{softmax}(r^e) 
\end{align}
where $\mathbf{W}^s_1, \mathbf{W}^s_2, \mathbf{W}^e_1, \mathbf{W}^e_2$ are trainable parameters of the projection function. 
$\mathbf{h}_s$ is the token representation of the start label. 
$p^s, p^e$ are the start and the end probability distributions over all tokens respectively, where $p^s \subseteq \mathbb{R}^L, p^e \subseteq \mathbb{R}^L$. 
Different from predicting start and end independently, we explicitly model the relation between them.
As shown in Equation~\ref{beam}, the calculation of end distribution depends on start position.
%
The training objective of token level prediction is
defined as the cross entropy loss of start and end
predictions:
\begin{eqnarray}
    \mathcal L_{t} = -\frac{1}{M} \sum_{j=1}^M[log(p^s_{y_j^s}) + log(p^e_{y_j^e})]
\end{eqnarray}
where $y_j^s$ and $y_j^e$ are the ground-truth of start and end positions of $j\mbox{-}$th example respectively. 
$M$ is the number of examples.

\noindent \textbf{Sentence level answer}. Encoder representation of $\mathtt{[CLS]}$ token $\mathbf{h}_{[\mathtt{CLS}]}$ is viewed as the sentence level feature, which is used to predict dialog acts and answerability by: 
\begin{align}
    p^{f} &= \delta(\mathbf{W}^f_2\mathtt{tanh}(\mathbf{W}^f_1\mathbf{h}_{[\mathtt{CLS}]})  \\
    p^{y} &= \delta(\mathbf{W}^y_2\mathtt{tanh}(\mathbf{W}^y_1\mathbf{h}_{[\mathtt{CLS}]}) \\
    p^{u} & = \sigma(\mathbf{W}^u_2\mathtt{tanh}(\mathbf{W}^u_1\mathbf{h}_{[\mathtt{CLS}]}) \label{unasnwer}
\end{align}
where $\mathbf{W}^f_1, \mathbf{W}^f_2, \mathbf{W}^y_1,\mathbf{W}^y_2, \mathbf{W}^u_1, \mathbf{W}^u_2$ are trainable parameters.
$\delta$ is a softmax function.
$\sigma$ is a sigmoid function.
$ p^f, p^y$ are prediction distributions of continuation and affirmation respectively, where $p^f \subseteq \mathbb{R}^3, p^y \subseteq \mathbb{R}^3$. 
$p^u$ is the prediction score of answerability. 
Their corresponding cross entropy losses are defined as:
\begin{align}
    \mathcal L_{ct} & = - \frac{1}{M}\sum_{j=1}^M[logp^f_{y_j^{ct}}]\\
    \mathcal L_{af} & = - \frac{1}{M}\sum_{j=1}^M[logp^y_{y_j^{af}}] \\
    \mathcal L_{na} & = - \frac{1}{M}\sum_{j=1}^M[y^{na}_jlogp^u + (1-y^{na}_j)log(1-p^u)] 
\end{align}
where $y_j^{ct}, y_j^{af}, y_j^{na}$ are the ground-truths of continuation, affirmation and answerability respectively.
The training objective of sentence level prediction is defined as:
\begin{align}
    \mathcal L_s = \mathcal L_{ct} + \mathcal L_{af} +\mathcal L_{na} 
\end{align}

\subsubsection{Answer Calibration}
The highest scoring span among the regional answers is sometimes not the span with the highest F1 score. 
Motivated by this issue, we introduce an answer calibration mechanism, with the goal of predicting more accurate regional answers. 
%
Particularly, given the answer candidates, we first compute their span representation, which is a weighted self-aligned vector:
\begin{align}
    \alpha^t & = \mathtt{softmax}(\mathbf{W}_r\mathbf{h}_{s_t:e_t}) \\
    \mathbf{c_t} & = \sum_{j=s_t}^{e_t} \alpha^t_j \mathbf{h}_j
\end{align}
where $\mathbf{W}_r$ is a trainable parameter. $\mathbf{h}_{s_t:e_t}$ is a shorthand for stacking a list of vectors $\mathbf{h}_j$ $(s_t \leqslant j \leqslant e_t)$. $s_t, e_t$ are the start and end of the $\mbox{$t$-th}$ answer candidate.
$\mathbf{c}_t$ is the span representation of the  $\mbox{$t$-th}$ answer candidate.
Then, all candidate representations are transferred to a multi-head self-attention layer ($\mathtt{Multi\mbox{-}SelfAtt}$) to capture the similarities and  the differences among them, and the detailed calculation is shown as: 
\begin{align}
      \mathbf{c}_t & = \mathbf{c}_t + \mathtt{Emb}(t)      \\
     \mathbf{c^{'}} & = \mathtt{Multi\mbox{-}SelfAtt}(\mathbf{c}) \\
     \beta & = \mathtt{softmax}(\mathtt{tanh}(\mathbf{W}_o \mathbf{c^{'}}_{0:T}))
\end{align}
where $\mathtt{Emb}$ is the position embedding of $t$ and 
a smaller value $t$ means a higher original prediction score, which is a valuable feature for the model to identify the different importance of candidates.
$\mathbf{c}$ is list of candidate representations, formally $\mathbf{c} = [\mathbf{c}_1,...,\mathbf{c}_T]$.
$\mathbf{W}_o$ is trainable parameter.
$\beta$ is the distribution of calibration score over the answer candidates, $\beta \subseteq \mathbb{R}^T$.
The cross entropy loss for answer calibration is given as:
\begin{equation}
    \mathcal L_{ac} = - \frac{1}{T}\sum_{j=1}^T[log\beta_{y_j^{ac}}]
\end{equation}
where $y_j^{ac}$ is a manually constructed label.
We define $y_j^{ac}$ is the candidate which obtains the highest F1 score with the gold span among all candidates.
If the highest F1 score is zero and the corresponding question is answerable, we randomly replace a candidate with the gold span.

\subsection{Document Reader}
%
The predicted spans from different chunks are compacted to a new text $P^{q}$ with a designed Minimum Span Coverage (MSC) algorithm, as shown in Algorithm~\ref{MSC}.
MSC guarantees that $P^{q}$ covers all regional spans and is sufficiently condensed to be encoded once.
\begin{algorithm}[htb]
\caption{Minimum Span Coverage} 
\label{MSC}
{\bf Input: } 
{\bf A} = $\{a_i\}^{NT}_{i=1}$ \\
{\bf A} is the set of regional spans from all chunks \\
$N$ is the number of chunks, $T$ is the number of regional spans for one chunk 
\begin{algorithmic}[1]
\For{$a_i$ in {\bf A}} 
    \For{{$a_{j,j\neq i}$ in {{\bf A}}}} 
        \If { $\mathtt{overlap}$($a_i$, $a_j$)}
            \State  {\bf A} += $\mathtt{coverage}$($a_i$, $a_j$) 
            \State  {\bf A} -= $a_i$, {\bf A} -= $a_j$
        \EndIf
    \EndFor
\EndFor
\State $\mathtt{coverage}$($a_i$,$a_j$) is the span corresponding to (start,end) = $(\mathtt{min}(s_i,s_j),\mathtt{max}(e_i,e_j))$, where $(s_i,e_i)$ and $(s_j,e_j)$ are (start,end) of $a_i$ and $a_j$.
\State Recursively execute the above steps until no condition of $\mathtt{overlap}$($a_i$,$a_j$)
\State Concatenate the elements in {\bf A} together as $P^q$
\State \Return $P^q$
\end{algorithmic}
\end{algorithm}

The input sequence of the text encoder is $\mathtt{X} = [\mathtt{[CLS]} ; q ; \mathtt{[SEP]} ; P^q]$, which is further read by the document reader to predict the answers as the global answers.
%
The span label of the document reader is the longest common substring between $P^{q}$ and the original gold span.

The global answers and the regional answers are aggregated as final predictions.
%
For answerability prediction, the document reader predicts a global no answer score $U_{g}$  and the chunk reader predicts a series of no answer score $U_r = \{u_k\}_{k=1}^N$, where $u_k$ is the predicted no answer score of $\mbox{$k$-th}$ chunk.
The final no answer score $\mathcal{S}_{na}$ is defined as:
\begin{align}
    \mathcal{S}_{na} & = \lambda U_{g} + (1-\lambda) \mathtt{min}(U_{r})  \label{unasnwerable}
\end{align}
where $\lambda$ is a hyperparameter to tune the weights of the global answers and the regional answers.

\subsection{Voting Strategy}
After aggregating the answer spans, we re-score them with a voting strategy that is based on a hypothesis: the spans predicted by both the chunk reader and the document reader are more likely to be correct. 
%
This strategy allows all spans to vote with each other to choose the most common span.
Concretely, the voting score of each span is obtained by: 
\begin{align}
    \mathtt{Voting}(x_i) & = \frac{1}{T\mbox{-}1}\sum_{j=1}^T(\mathtt{F1}(x_i,x_{j,j \neq i})) \label{cross} \\
    \mathtt{F1}(x_i,x_j) & = \frac{2R(i,j)P(i,j)}{R(i,j)+P(i,j)} \\
    R(i,j) & = \frac{|x_i\cap x_j|}{|x_j|} \\
    P(i,j) & =  \frac{|x_i\cap x_j|}{|x_i|}
\end{align}
where $| \ |$ denotes the number of words, $|x_i\cap x_j|$ denotes the number of the common words between $x_i$ and $x_j$.
The function of $\mathtt{F1}(x_i,x_j)$ represents the sequence similarity between $x_i$ and $x_j$.  
A larger voting score means that the corresponding span is similar to more candidates than the others. 
%
Finally, the voting strategy reranks answer spans according to the original prediction score $\mathcal{S}(x)$ and the voting score: 
\begin{align}
    \label{final_score}
    \mathtt{score}(x) = \gamma \mathcal{S}(x) + (1-\gamma)\mathtt{Voting}(x)
\end{align}
$\gamma$ is the weight of two scores.

\subsection{Training and Inference}
We adopt the multi-task learning idea to jointly learn the predictions of answer span, answerability and dialog acts.
All parameters are trained with an end-to-end manner. 
The training loss of the chunk reader $L_{c}$ and the document reader $L_{d}$ are:
\begin{align}
    \mathcal L_{c} &= \mathcal L_t +  \mathcal L_{s} +  \mathcal L_{ac} \\
    \mathcal L_{d} & =  \mathcal L_t +  \mathcal L_{s}
\end{align}

%
The detailed training and inference processes are given in Algorithm~\ref{process}.
\begin{algorithm}[htb]
\caption{Training and Inference Process} 
\label{process}
{\bf Input: } 
{\bf D} = training set, {\bf d} = test set \\
{\bf Initialize: }$\Theta_1, \Theta_2$ $\leftarrow$ pre-trained parameters 
\begin{algorithmic}[1]
\State {\bf Training}
\State Train $\Theta_1$ on {\bf D} with $\mathcal L_c$, then predict answers on {\bf D} to construct a new dataset {\bf D$^{'}$} through Algorithm~\ref{MSC}
\State Train $\Theta_2$ on {\bf D$^{'}$} with $ \mathcal L_d$
\State {\bf Inference}
\State $\Theta_1$ predict the regional answers set $A$ on {\bf d} and construct a new dataset {\bf d$^{'}$} through Algorithm~\ref{MSC}
\State $\Theta_2$ predict the global answers set $A^{'}$ on {\bf d$^{'}$}
\State Aggregate and rerank the final answers set $\widetilde{A} = A \cup A^{'}$ with Equation~\ref{final_score}
\end{algorithmic}
\end{algorithm}


\subsection{Experiment Setup}
\subsubsection{Dataset}
Our experiments are mainly conducted on two long document datasets QuAC (Question Answering in Context)~\cite{quac} and TriviaQA~\cite{triviaqa}.
QuAC is a large-scale dataset created for simulating information-seeking conversations. 
Its questions are often more open-ended, unanswerable, or only meaningful within the dialog context.
TriviaQA is a large-scale open-domain MRC dataset, which requires cross sentence reasoning to find answers. 
It contains data from Wikipedia and Web domains, where Wikipedia subset is used in our work.
%
The statistic information of these two dataset is summarized in Table~\ref{Dataset}.

\begin{table}[H]
\renewcommand{\arraystretch}{1.5} 
  \centering
  \fontsize{10}{10}\selectfont
    \begin{tabular}{|l | c c c|}
    \hline
    ~&\textbf{Train}&\textbf{Dev}&\textbf{Test}\cr
    \hline
    \textbf{TriviaQA} &~&~&~\cr
    \# questions&61,888&7,993&7,701 \cr
    \# tokens / input &11,222&11,382&-\cr
    \hline
    \textbf{QuAC} &~&~&~\cr
    \# questions&83,568&7,354&7,353 \cr
    \# tokens / input &641 & 707 &-\cr
    \# dialogs&11,567&1000&1002 \cr
    \# question / dialog &7.2&7.4&7.4 \cr
    \% unanswerable &20.2&20.2&20.1 \cr
    \hline
 \end{tabular}
  \caption{Statistics of two datasets. \# denote the number of each item. \% denote a percentage value. }
  \label{Dataset}
\end{table}

\subsubsection{Evaluation Metrics}
For answer span prediction,  the QuAC challenge provides two
evaluation metrics, the word-level F1 and the human equivalence
score (HEQ). 
The word-level F1 measures the overlap of the prediction and the gold span after removing stopwords. 
%
HEQ measures the percentage
of examples for which model F1 score is higher than the average human F1 score.
HEQ contains two variants HEQ-Q and HEQ-D.
HEQ-Q is 1 if model performance exceeds the human performance for each question.
HEQ-D is 1 if model performance of all the questions
in the dialog exceeds human.
For dialog act prediction, the accuracy is adopted as evaluation metric.
For the TriviaQA dataset, word-level F1 score and exact match (EM) are used as evaluation metrics.

\subsubsection{Implementation Details}
We tried three different PLMs, BERT-large~\footnote{https://github.com/google-research/BERT}, ELECTRA-large~\footnote{https://github.com/google-research/electra} and Longformer-large~\footnote{https://github.com/allenai/Longformer} as initialization parameters of text encoder to verify the effectiveness of RoR comprehensively.  
%
The max sequence length of questions is set to 128 and the answer length is set to 64.
The stride of the sliding window for splitting documents is set to 128.
The batch size is set to 12.
%
The model is optimized using Adam~\cite{adam} with learning rate = 2e-5, maximal gradient clipping = 1.0. 
The hyperparameter $\lambda$ is set to 0.9, $\gamma$ is set to 0.5.
%
In the inference process, we use beam search to predict end position based on start position and the beam size is 5.
The decision of answerability depends on the numerical comparison between the no answer score $\mathcal S_{na}$ in equation~\ref{unasnwerable} and a threshold $\zeta$, which is set to 0.3.
If  $\mathcal S_{na}$ is higher than $\zeta$, the corresponding question is unanswerable.

\begin{table*}
\renewcommand{\arraystretch}{1.5} 
  \centering
  \fontsize{9}{10}\selectfont
    \begin{tabular}{|l| c c c c c c|}
    \hline
    
    \textbf{Model}&{\textbf{F1}}&{\textbf{HEQ-Q}}&{\textbf{HEQ-D}}&{\textbf{Answerability}}&{\textbf{Continuation}}&{\textbf{Affirmation}}\cr \hline 
    Longformer &74.3 & 71.5& 14.5&76.5 & 64.4 & 89.8  \cr \hline
    BERT &67.4 & 63.7& 7.9 &68.3&63.3&88.4\cr
    BERT-RoR &\textbf{69.6} & \textbf{65.8}& \textbf{9.8} &\textbf{74.7}&\textbf{63.5}&\textbf{88.6}\cr \hline
    ELECTRA&73.8&71.2&14.3&76.1&64.9&89.7\cr
    ELECTRA-RoR &\textbf{75.7}&\textbf{73.4}&\textbf{17.8}&\textbf{78.2}&\textbf{65.0}&\textbf{90.0}\cr \hline
 \end{tabular}
  \caption{Results on the development set of QuAC.}
  \label{quac result}
\end{table*}

MSC algorithm is important in RoR which guarantees the condensed form of an arbitrarily long input sequence shorter than 512 tokens through limiting the total number and length of the regional answers.
In practice, we perform the following four operations on TriviaQA and QuAC:
\begin{itemize}
    \item The max number of answer candidates for each chunk $T$ is set to 5.
    \item A document is split up to 7 chunks in QuAC and 15 chunks in TriviaQA.
    \item The regional answer is truncated if longer than 15 tokens.
    \item Many regional answers overlap or even differ by a few words. MSC algorithm removes the duplicate words to ensure the condensed document shorter.
\end{itemize}
After the operations above, the longest condensed document contains 471 tokens in TriviaQA and 184 tokens in QuAC. When RoR is adapted to other datasets, the length of the condensed documents can be guaranteed to be shorter than 512 as long as the parameters in the above four operations are adjusted correspondingly.

In order to improve the model performance, some data augmentations are applied to better train the model. 
Specifically, ELECTRA is fine-tuned on other MRC datasets before fine-tuned on QuAC, such as SQuAD~\cite{squad2} and CoQA~\cite{coqa}, hoping to transfer the knowledge in other datasets to our model. 
Experimental results show that CoQA has a much higher lifting effect than SQuAD. 
This is because both CoQA and QuAC are conversational MRC datasets, while SQuAD is a single-turn MRC dataset.
The answers in CoQA are free-form and generally short (average answer length = 2.7), which is quite different from QuAC (average answer length = 15.1).
As a result, we choose the rationale sentence of the gold span in CoQA as the prediction target. 

\subsection{Main Results}
\textbf{Results on QuAC}. Table~\ref{quac result} displays the experimental results on the development set of QuAC. 
In view of the fact that the sequence length in QuAC dataset does not exceed the encoding length limit of Longformer (\textit{i.e.,} 4096), we did not apply RoR to the Longformer.
The results show that RoR significantly improves the performance of PLMs on all  evaluation metrics, illustrating the effectiveness of RoR on long document modeling.
Among the three PLMs, Longformer, which can encode longer sequences, performs best, followed by ELECTRA.
Nevertheless, with the enhancement of RoR, ELECTRA-RoR outperforms Longformer and achieves state-of-the-art results over all metrics on the dev set of QuAC.

\begin{table}[H]
\renewcommand{\arraystretch}{1.5} 
  \centering
  \fontsize{8}{9}\selectfont
  \vspace{0.1cm}
    \begin{tabular}{|l | c c c|}
    \hline
    \textbf{Model}&\textbf{F1}&\textbf{HEQ-Q}&\textbf{HEQ-D}\cr
    \hline
    \hline
    Human &81.1&100&100\cr \hline 
    \hline
    ELECTRA-RoR &\textbf{74.9}&\textbf{72.2}&\textbf{16.4} \cr\hline
    \hline
    EL-QA &74.6&71.6&16.3 \cr
    History QA &74.2&71.5&13.9 \cr 
    TR-MT &74.4&71.3&13.6 \cr  
    GraphFlow~\cite{graph_flow} &64.9&60.3&5.1 \cr 
    HAM~\cite{ham} &65.4&61.8&6.7 \cr 
    FlowDelta~\cite{flow_delta} &65.5&61.0&6.9 \cr 
    HAE~\cite{hae} &62.4&57.8&5.1 \cr 
    FlowQA~\cite{flowqa} &64.1&59.6&5.8 \cr 
    BiDAF++ \textit{w/ 2-Context} &60.1&54.8&4.0 \cr    
    BiDAF++~\cite{bidaf++} &50.2&43.3&2.2 \cr 
    \hline

%
 \end{tabular}
  \caption{Test results on QuAC with sample methods on the leaderboard \url{https://quac.ai/}.}
  \label{test result}
\end{table}

\begin{table*}
\renewcommand{\arraystretch}{1.5} 
  \centering
  \fontsize{10}{10}\selectfont
    \begin{tabular}{|l | c c c c c c |}
    \hline
    \textbf{Model}&\textbf{F1}&\textbf{HEQ-Q}&\textbf{HEQ-D}&\textbf{Answerability}&\textbf{Continuation}&\textbf{Affirmation}\cr
    \hline
    chunk reader &73.8&71.2&14.3&76.1&64.9&89.7\cr
     w/ calibration &74.1&71.4&15.3&76.9&64.6&89.7\cr
     \ \ w/ document reader &74.8&72.1&15.7&77.4&64.7&89.7\cr
     \ \ \ \ w/ voting strategy &75.4&72.9&16.9&77.4&64.7&89.7\cr
     \ \ \ \ \ \  w/ knowledge transfer &75.7&73.4&17.8&78.2&65.0&90.0\cr
    \hline

 \end{tabular}
  \caption{Ablation study of ELECTRA-RoR on the development set of QuAC.}
  \label{ablation}
\end{table*}

\noindent \textbf{Official leaderboard results on QuAC}. QuAC challenge provides a hidden test set, where the dialog acts prediction is not the main task of the leaderboard and their evaluation scores are not considered in the final model ranking.
Table~\ref{test result} displays the span prediction results of all baselines and our model, from which we can see that our model ELECTRA-RoR outperforms the previous best performing model EL-QA and achieves new state-of-the-art on all three metrics.
From the results of the leaderboard, we observe that the top ranking models almost all use the advanced pre-trained models, such as ELECTRA, RoBERTa and BERT.
Although some models have a well-designed model structure, they still lag behind the models that uses the pre-trained models as encoder, showing the powerful modeling capabilities of the pre-trained model.
For example, FlowDelta boosts the F1 score of FlowQA from 64.1 to 65.5 with the help of BERT.
Compared to BiDAF++, BiDAF++ \textit{w/ 2-Context} incorporates two turns of previous dialog history and significantly improves the performance of BiDAF++, verifying the importance of historical information in the conversational MRC. 

\noindent \textbf{Results on TriviaQA}.
Table~\ref{result} displays the experimental results on the TriviaQA dataset.
The experimental results show that RoR is able to comprehensively improve the score of the PLMs and the average F1 gain is 2.1, proving that RoR is also highly effective in the task of open-domain MRC.

%
Similar to the performance on the QuAC dataset, among the three PLMs, Longformer performs best on the TriviaQA dataset, ELECTRA followed, and the worst is BERT.
Although Longformer has such excellent performance which used to be state-of-the-art model on the TriviaQA dataset, RoR still further improve its performance (F1 score from 77.8 to 80.0).
The test results show that all improvements of RoR in Table~\ref{test result} and Table~\ref{result} are statistically significant (paired t-test, p-value < 0.01).

\begin{table}[H]
\renewcommand{\arraystretch}{1.5} 
  \centering
  \fontsize{9}{10}\selectfont
    \begin{tabular}{|l | c c|}
    \hline
    \textbf{Model}&\textbf{F1}& \textbf{EM}\cr
    \hline
    BERT& 68.4& 60.7 \cr
    BERT-RoR& \textbf{70.3} & \textbf{62.1}\cr \hline
    ELECTRA& 70.6 & 65.3\cr
    ELECTRA-RoR& \textbf{72.9} & \textbf{67.8}\cr \hline
    Longformer& 77.8 & 73.0\cr 
    Longformer-RoR&\textbf{80.0}&\textbf{75.0}
    \cr \hline 
 \end{tabular}
  \caption{Results on the TriviaQA dataset.}
  \label{result}
\end{table}

\subsection{Ablation Test}

We conduct an ablation analysis on development
set of QuAC to investigate the contributions of each module of the best model ELECTRA-RoR.

Table~\ref{ablation} displays the results of ablated systems, where we gradually add the proposed modules to the model structure.
It can be observed that answer calibration mechanism boosts the performance of the chunk reader in all three span evaluation metrics.
An interesting phenomenon is the calibration mechanism has the ability to improve the prediction accuracy of unanswerable questions.
This may be because we mask some training instances during calculating the calibration loss $\mathcal L_{ac}$ if their questions are unanswerable, since all span candidates are not correct.
This may provide some supervision signals for model training to identify unanswerable questions.
Next we analyze the contributions of the document reader. 
It can be seen that the evaluation scores of span prediction are further improved  when adding the document reader.
Especially the F1 score is improved by 0.7.
Moreover, the accuracy of answerability prediction is also improved.
This is because the document reader predicts a global no answer score which contributes to the final decision of answerability through equation~\ref{unasnwerable}.
Afterwards, the ablation results show that the voting strategy yields substantial improvement over the RoR model.
%
Finally, we can see that the transfer module could also comprehensively enhance the model performance, proving the efficiency of knowledge transfer in MRC task.

In terms of continuity and affirmation, the accuracy scores do not change much in the ablation experiments.
Nevertheless, we cannot completely ignore them when training our model, as we find that the performance of RoR will drop if remove the losses of $\mathcal L_{ct}$ and $\mathcal L_{af}$.
This reflects that the dialog acts can indeed encourage the encoder to produce more generic representations that benefit the answer span prediction task.

\subsection{Discussion on Answer Calibration}
As shown in Table~\ref{ablation}, the answer calibration mechanism improves the performance of the chunk reader. 
This section further discusses its other promoting effects on RoR.
The predicted regional answers determine the quality of global answers that will be predicted by the document reader. 
An ablation test on the condensed document $P^q$ is conducted on the dev set.
We find that the average F1 score of $P^q$ drops from 30.3 to 29.7 if without the answer calibration.
This test validates that the calibration mechanism is able improve the input accuracy of the document reader.
Meanwhile, the quality of global answers are further improved.
%

We also attempt to integrate the answer calibration mechanism into the document reader, but the results show that the improvement is not significant.
We speculate the reason is that the predictions of the document reader is sufficiently accurate.

%
%
%
%

\begin{figure}
\centering
\includegraphics[width=7.5cm]{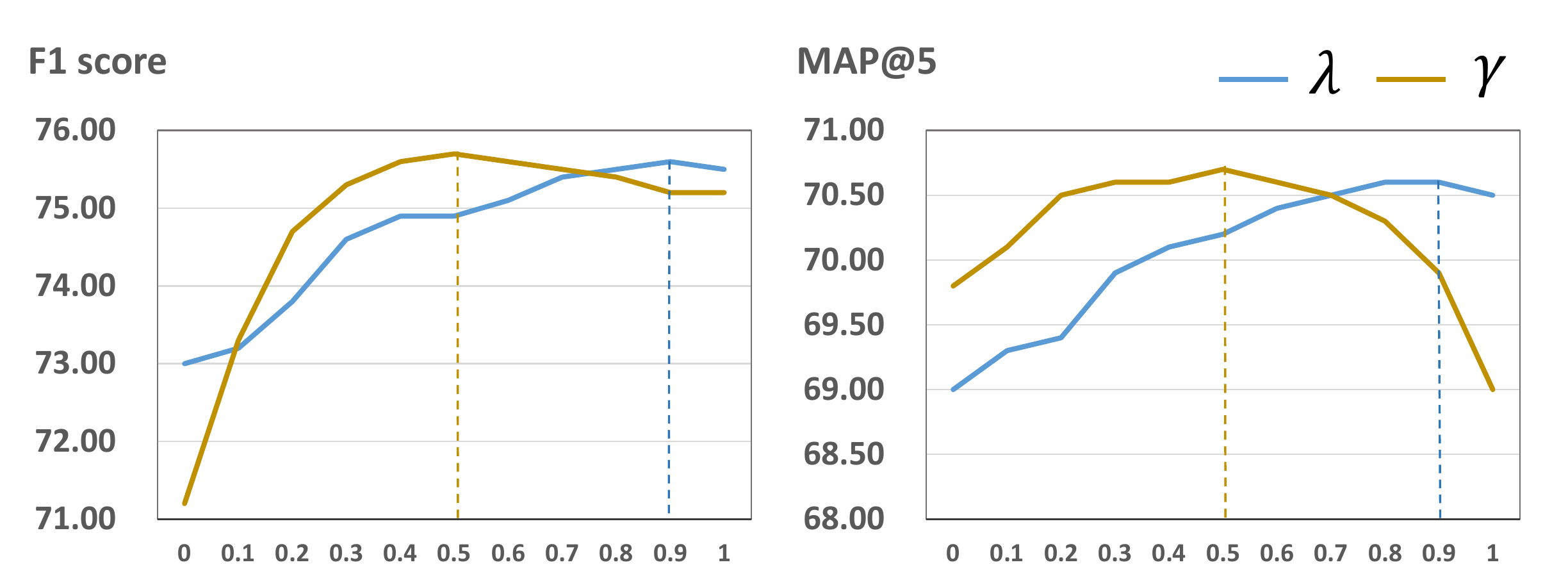}
\caption{The influence of weight.}
\label{weight}
\end{figure}

\subsection{Influence of Weight Parameter}
In the voting strategy, the parameter $\gamma$ weights the prediction score and the voting score.
In the process of answers aggregation,  $\lambda$ weights the score of the regional answers and the global answers. 
This section explores the influence of these two parameters on the model.

Figure 4 displays the F1 and mean average precision (MAP) curves of ELECTRA-RoR on QuAC with respect to different $\lambda$ and $\gamma$.
The results suggest that the model performance are sensitive to the weights. 
As the weights increases, both curves show a trend of increasing at first and then decreasing, reaching peaks at 0.5 and 0.9 respectively.
For the changing degree of the curves,  $\gamma$ actually has a greater influence on RoR than $\lambda$.
We notice that MAP score dramatically declines when reducing the weight of the voting score (\textit{i.e.,} $\gamma$ from 0.5 to 1.0), indicating that the voting score is a reliable basis to rerank the final prediction results.


\section{Conclusion}
In this work, a \textit{read-over-read} (RoR) pipeline is proposed for long document MRC, which contains an enhanced chunk reader to predict the regional answers and a document reader to predict the global answers.
Moreover, a voting strategy is designed to optimize the process of answer aggregation in RoR.
Comprehensive empirical studies on QuAC and TriviaQA demonstrate the effectiveness of RoR, which comprehensively improves the performances of the PLMs on long document reading. Meanwhile, ELECTRA-RoR achieves state-of-the-art over all evaluation metrics on QuAC leaderboard.

\section{Acknowledge}
We would like to thank three anonymous reviewers for their useful feedback. We also sincerely appreciate He He and Mark Yatskar for their help in evaluating our models on QuAC dataset.
This work is supported by the National Key Research and Development Program of China under
Grant No. 2020AAA0108600.

\bibliography{anthology,ror_new}
\bibliographystyle{acl_natbib}

\end{document}